# Ridge Estimation-Based Vision and Laser Ranging Fusion Localization Method for UAVs

**Huayu Huang,[1, 2,†] Chen Chen,[1,2,†] Banglei Guan,[1,2,*] Ze Tan,[1,2] Yang Shang,[1,2] Zhang Li,[1,2] and Qifeng Yu[1,2]**

[1]*College of Aerospace Science and Engineering, National University of Defense Technology, Changsha 410073, Hunan, China*
[2]*Hunan Provincial Key Laboratory of Image Measurement and Vision Navigation, Changsha 410073, Hunan, China*
[†]*These authors contributed equally.*
**guanbanglei12@nudt.edu.cn*

**Abstract:** Tracking and measuring targets using a variety of sensors mounted on UAVs is an effective means to quickly and accurately locate the target. This paper proposes a fusion localization method based on ridge estimation, combining the advantages of rich scene information from sequential imagery with the high precision of laser ranging to enhance localization accuracy. Under limited conditions such as long distances, small intersection angles, and large inclination angles, the column vectors of the design matrix have serious multicollinearity when using the least squares estimation algorithm. The multicollinearity will lead to ill-conditioned problems, resulting in significant instability and low robustness. Ridge estimation is introduced to mitigate the serious multicollinearity under the condition of limited observation. Experimental results demonstrate that our method achieves higher localization accuracy compared to ground localization algorithms based on single information. Moreover, the introduction of ridge estimation effectively enhances the robustness, particularly under limited observation conditions.

## 1. Introduction

UAVs stand out because of their many advantages, such as economy, remote control, and no casualties. It is widely used in smart cities [1,2], obstacle detection [3,4], harbor-border inspection [5,6], object recognition [7,8], and other fields [9–11]. Almost all of these applications require UAVs to track and locate the fixed target of interest quickly and accurately by carrying various sensor devices on them. Therefore, the algorithm of utilizing various sensor data from UAVs for rapid and precise localization of ground targets is of vital importance.

Traditional ground localization methods mostly combine distance, angle, visible light, or SAR images data of a single sensor with platform position information. However, each sensor has its specific shortcomings. For instance, monocular cameras have low measurement accuracy in the visual axis direction and stereo camera's baseline is insufficient for long distance measurement. Laser-based ranging localization methods rely on multiple consecutive observations and it is greatly influenced by lighting conditions and the target material. Using a single sensor alone can result in low accuracy and instability. Using certain novel sensors can address some challenging conditions. For instance, targets move so fast that the images are blurred; challenging illumination conditions result in rapid changes in lighting during a shot [12]. However, it can't address the issue of instability of the algorithm. To address the issue of low estimation accuracy and instability by a single sensor, many scholars have proposed using multiple information for fusion localization to make up for the shortcomings of a certain sensor [13–15].

Wang et al. [16] proposed a navigation system that integrates monocular vision with inertial measurement. By fusing the data from both sensors using an Extended Kalman Filter (EKF), the system avoids the accumulation of errors that arise when position and attitude are estimated separately by the two sensors. Li et al. [17] proposed an indoor multi-sensor fusion localization method based on federated filtering, aimed at enhancing the indoor localization accuracy of mobile robots. This approach integrates data from LiDAR, visual odometry, and IMU through the design of a hybrid federated filter to achieve dynamic sensor data fusion. Their method demonstrates high localization accuracy and robustness in indoor environments. Ren et al. [18] combined the visual data from UAV with satellite remote sensing data that incorporates geographic location information. By matching the images captured by the UAV with satellite remote sensing images, they were able to locate the targets captured by the UAV. To avoid the problem that the visible light camera fails to work, Li et al. [19] combined the information of inertial measurement unit (IMU) and Ultra-Wideband (UWB) to conduct cooperative navigation for UAV swarms in the case of GNSS-denied.

One of the most common applications of multi-sensor fusion is the combination of a visible light camera and laser range finder. Zhang et al. [20] presented a sensor fusion strategy of monocular cameras and laser rangefinders applied for SLAM in dynamic environments. This method incorporates two individual Extended Kalman Filter (EKF) based SLAM algorithms: monocular and laser SLAM. The error of localization in fused SLAM is reduced compared with those of individual SLAM. Chan et al. [21] presented a novel method for laser SLAM and visual SLAM fusion which provided robust localization. Instead of using feature matching methods to achieve the fusion procedure, trajectories matching is proposed with an attempt to achieve the generalization over all different kinds of SLAM algorithms. This architecture can be applied to fusing any two kinds of laser-based SLAM and monocular camera-based SLAM together.

Chao et al. [22] analyzed the reason why the measurement error of the monocular camera in the visual axis direction is much larger than that in the vertical direction. They proposed that the overall performance of the measurement system could be improved by incorporating a laser rangefinder for auxiliary measurements. However, they only utilized a single observation data point from the monocular camera and laser rangefinder, resulting in insufficient measurement accuracy and robustness under extreme observation conditions. Aimed at improving the low measurement accuracy of the binocular vision sensor along the optical axis, Wang et al. [23] proposed a method of utilizing laser ranging sensors for auxiliary calibration. They employed federated filtering to improve the information utilization and measurement accuracy. The drawback of this method is that the binocular vision sensor is constrained by the baseline length, resulting in insufficient measurement accuracy under limited observation conditions. Xiong et al. [24] proposed a lightweight relative localization system for micro-UAV swarms. This system is predicated on the fusion of UWB, visual, and IMU data, implemented through distributed graph optimization. The system achieves high-precision relative localization in the absence of GNSS support. Notwithstanding the adoption of distributed optimization techniques, this approach still requires substantial computational resources to fuse and process data from multiple sensors in real-time. Zhang et al. [25] proposed a multimodal fusion-based method for the accurate localization of static small targets against uniform backgrounds. This approach achieved accurate detection and three-dimensional spatial localization by fusing the data from CMOS cameras, laser rangefinders, and angle sensors. The limitation of this methodology is the complexity of its multimodal data fusion and processing method, which necessitates substantial computational resources. Moreover, this work primarily addresses static small targets against uniform backgrounds, thus exhibiting limited generalization.

In the actual observation scene, observing in an area far away from the target is usually necessary. Compared with the downward-looking observation at the same flight height, the observation angle of the long-distance observation is larger, and the angle between the two observation lines of sight is smaller, which can easily lead to limited observation conditions such as small intersection angle and large inclination angle. At this time, when using the least squares estimation algorithm, the column vectors of the design matrix have serious multicollinearity [26]. Under such limited observation conditions, current multi-sensor fusion localization methods suffer from significant instability and low robustness due to the serious multicollinearity in their design matrices. To address the problem of insufficient localization accuracy when using a monocular camera or laser range finder alone under limited observation conditions, this paper proposes a fusion approach that integrates and normalizes the observational data from monocular camera and laser range finder, significantly enhancing localization precision in limited environments. Furthermore, to mitigate the serious multicollinearity in the design matrix, this paper proposes a least squares localization algorithm based on ridge estimation, integrating vision and ranging fusion. To validate the superiority of the proposed algorithm in this paper, we conducted simulation experiments and flight experiments. Experimental results show that the introduction of ridge estimation effectively enhances the accuracy and robustness of the algorithm, especially under limited observation conditions.

The main contributions of this paper include:

- This paper presents a novel localization algorithm that fuses visual and laser ranging data to improve target localization accuracy. This fusion approach leverages the complementary strengths of both sensors to transcend the limitations of traditional single sensor.
- The proposed method fuses visual and ranging data and normalizes the different data to ensure consistency in physical units. This normalization enables the simultaneous utilization of the advantages of both vision and laser ranging.
- To address multicollinearity issues under limited observation conditions, this paper introduces ridge estimation into the least squares localization algorithm. By adding a penalty function, the algorithm based on ridge estimation achieves more robust and accurate estimation.

## 2. Vision and ranging fusion algorithm based on ridge estimation

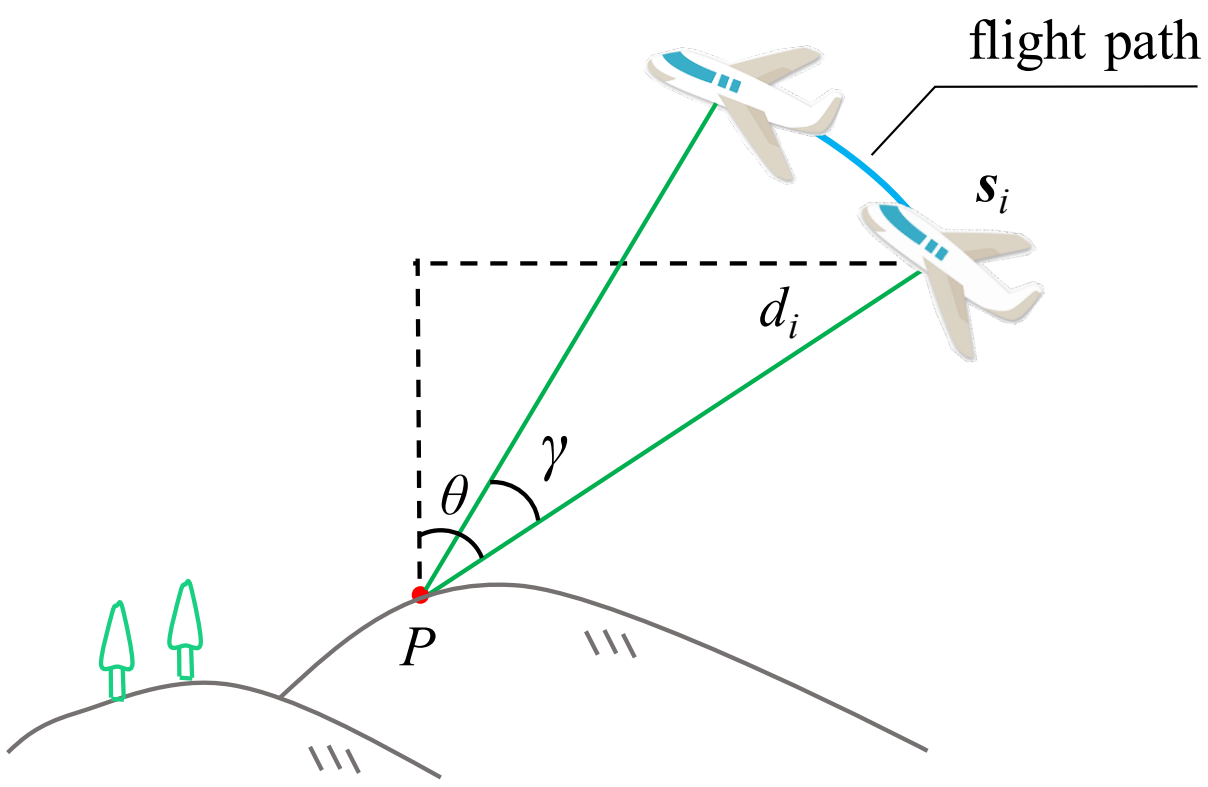

Fig. 1. Ground target localization with an airborne electro-optical platform.

As shown in Fig.1, the airborne photoelectric flight platform performs $n$ consecutive observations on the target. The angle $\theta$ between the observation line of sight and the vertical direction is the observation angle, and the angle $\gamma$ between the first and last observation lines is the intersection angle. The platform position of the $i$ -th ($i \leq n$) observation is $s_i$. The laser range finder is used to record the distance information $d_i$ between the platform and the target, and the target is locked in the center of the field of view and imaged by a monocular camera. Usually, observation errors are introduced in the measurement process, so the least square method can be used to minimize the sum of squares of errors.

However, in actual measurements, there are often limited observation conditions, such as small intersection angles and large inclination angles. The small intersection angle condition typically refers to the scenario in which the angle $\gamma$ is less than 30 °. Under such condition, the same line-of-sight angular error will result in more significant positioning errors. On the other hand, the large inclination angle condition is shown in Fig.2. Generally, the downward-looking is the optimal observation, as shown in Fig.2(a). However, in complex three-dimensional environments, observations of distant target areas often need to be conducted at a certain altitude. This will lead to the inclination angle, as shown in Fig.2(b). As the observation distance increases, the observation inclination angle becomes larger. As shown in Fig.2(c), under a large inclination angle, the geometric shape of the image undergoes significant distortion, and the angular error of the line of sight increases notably.

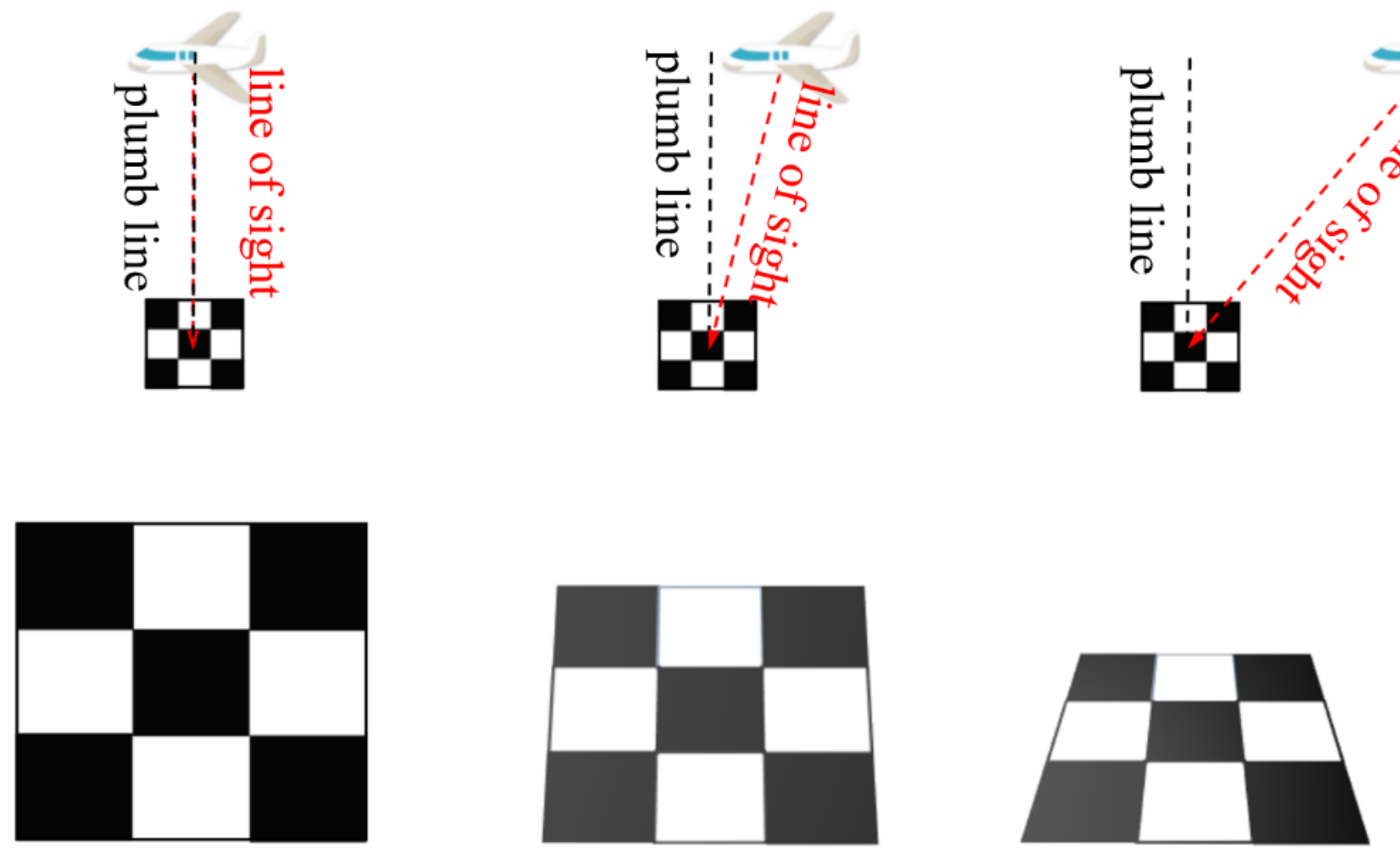

(a) The downward-looking (b) A small inclination angle (c) A large inclination angle

Fig. 2. Imaging maps of different inclination angles.

This section proposes the fusion localization method based on ridge estimation to address the problems of limited observation conditions. First, Section 2.1 introduces the least squares algorithm based on visual information. Second, Section 2.2 introduces the least squares algorithm based on laser ranging. Then, Section 2.3 proposes the least squares algorithm of the fusion method and introduces the normalization of the different data. Finally, Section 2.4 introduces ridge estimation to the least squares algorithm to address the problems of limited observation conditions.

### *2.1 Least squares algorithm based on visual information*

When the monocular camera is fixedly installed with the IMU, the relative rotation angle of the camera during the movement is the same as that of the IMU. At this time, the relative pose of the moving camera can be estimated with at least four homonymous points to solve the projection matrix $\boldsymbol{M}$ of the target from the earth coordinate system to the image coordinate system [27]. Therefore, the three-dimensional coordinates of the ground target can be solved by using at least two frames of images from the monocular camera. However, due to measurement errors, we usually combine redundant observations to minimize the sum of squares of the observation errors.

Assuming that the three-dimensional coordinate of a point in the earth coordinate system is $(X, Y, Z)$, the coordinate in the camera coordinate system at a certain moment is $(X_c, Y_c, Z_c)$, and the pixel coordinate corresponding to the image point $p_i$ on a frame image is $(\tilde{x}_i, \tilde{y}_i)$, the relationship between the three-dimensional space point and the two-dimensional image point can be expressed as:

$$Z_c \begin{bmatrix} \tilde{x}_i \\ \tilde{y}_i \\ 1 \end{bmatrix} = \boldsymbol{M}_i \begin{bmatrix} X \\ Y \\ Z \\ 1 \end{bmatrix}, \tag{1}$$

where $\boldsymbol{M}$ is the projection matrix, which can be expressed by the rotation matrix $\boldsymbol{R}_i$ and the translation vector $\boldsymbol{T}_i$ from the earth coordinate system to the camera coordinate system and the parameter matrix $\boldsymbol{K}$ in the camera:

$$\boldsymbol{M}_i = \boldsymbol{K} \begin{bmatrix} \boldsymbol{R}_i & \boldsymbol{T}_i \\ \boldsymbol{0}^T & 1 \end{bmatrix} = \begin{bmatrix} m_0^i & m_1^i & m_2^i & m_3^i \\ m_4^i & m_5^i & m_6^i & m_7^i \\ m_8^i & m_9^i & m_{10}^i & m_{11}^i \end{bmatrix}. \tag{2}$$

Assuming that the projection matrix $\boldsymbol{M}$ of the camera is known, where the intrinsic parameters can be calibrated in advance and the pose can be obtained through IMU and GPS [27], or the technique of structure from motion within a given scene [28-30]. Considering that the target satisfies $Z_c \neq 0$ in the camera coordinate system, the pixel value can be expressed as:

$$\begin{cases} \tilde{x}_i = \dfrac{m_0^i X + m_1^i Y + m_2^i Z + m_3^i}{m_8^i X + m_9^i Y + m_{10}^i Z + m_{11}^i} \\ \tilde{y}_i = \dfrac{m_4^i X + m_5^i Y + m_6^i Z + m_7^i}{m_8^i X + m_9^i Y + m_{10}^i Z + m_{11}^i} \end{cases}. \tag{3}$$

The error model of the above formula is expressed as:

$$\boldsymbol{\psi} = F(\boldsymbol{X}) + \boldsymbol{\Delta}, \tag{4}$$

where $\boldsymbol{\psi}$ is the observed value, $\boldsymbol{\Delta}$ is the observation error, and $F(\boldsymbol{X})$ is the nonlinear function of the independent variable $\boldsymbol{X} = [X \quad Y \quad Z]^T$:

$$\boldsymbol{\psi}_{2n\times1} = [\tilde{x}_1 \quad \tilde{y}_1 \quad \cdots \quad \tilde{x}_n \quad \tilde{y}_n]^T, \tag{5}$$

$$F(\boldsymbol{X}) = \begin{bmatrix} \dfrac{m_0^1 X + m_1^1 Y + m_2^1 Z + m_3^1}{m_8^1 X + m_9^1 Y + m_{10}^1 Z + m_{11}^1} \\ \dfrac{m_4^1 X + m_5^1 Y + m_6^1 Z + m_7^1}{m_8^1 X + m_9^1 Y + m_{10}^1 Z + m_{11}^1} \\ \vdots \\ \dfrac{m_0^n X + m_1^n Y + m_2^n Z + m_3^n}{m_8^n X + m_9^n Y + m_{10}^n Z + m_{11}^n} \\ \dfrac{m_4^n X + m_5^n Y + m_6^n Z + m_7^n}{m_8^n X + m_9^n Y + m_{10}^n Z + m_{11}^n} \end{bmatrix}_{2n\times1}, \tag{6}$$

$$\boldsymbol{\Delta}_{2n\times1} = [\Delta_{\tilde{x}_1} \quad \Delta_{\tilde{y}_1} \quad \cdots \quad \Delta_{\tilde{x}_n} \quad \Delta_{\tilde{y}_n}]^T. \tag{7}$$

For ease of numerical computation, the expression for $\boldsymbol{M}_i$ can be written as:

$$\begin{cases} M_1^i = m_0^i X + m_1^i Y + m_2^i Z + m_3^i \\ M_2^i = m_4^i X + m_5^i Y + m_6^i Z + m_7^i \\ M_3^i = m_8^i X + m_9^i Y + m_{10}^i Z + m_{11}^i \end{cases}. \tag{8}$$

We use the result of the forward intersection of $n$ frames of images [27] as the initial value $\boldsymbol{X}_0$, and carry out the first-order Taylor expansion of $F(\boldsymbol{X})$ at $\boldsymbol{X}_0$, then:

$$F(\boldsymbol{X}) = F(\boldsymbol{X}_0) + \boldsymbol{H}d\boldsymbol{X} + \boldsymbol{\Delta}, \tag{9}$$

where $\boldsymbol{H}$ is the design matrix, $d\boldsymbol{X}$ is the correction value of the initial value $\boldsymbol{X}_0$:

$$\boldsymbol{H}_{2n\times 3} = \begin{bmatrix} \frac{m_0^1}{M_3^1} - \frac{m_8^1 M_1^1}{(M_3^1)^2} & \frac{m_1^1}{M_3^1} - \frac{m_9^1 M_1^1}{(M_3^1)^2} & \frac{m_2^1}{M_3^1} - \frac{m_{10}^1 M_1^1}{(M_3^1)^2} \\ \frac{m_4^1}{M_3^1} - \frac{m_8^1 M_1^1}{(M_3^1)^2} & \frac{m_2^1}{M_3^1} - \frac{m_{10}^1 M_1^1}{(M_3^1)^2} & \frac{m_6^1}{M_3^1} - \frac{m_{10}^1 M_1^1}{(M_3^1)^2} \\ \vdots & \vdots & \vdots \\ \frac{m_0^n}{M_3^n} - \frac{m_8^n M_1^n}{(M_3^n)^2} & \frac{m_1^n}{M_3^n} - \frac{m_9^n M_1^n}{(M_3^n)^2} & \frac{m_2^n}{M_3^n} - \frac{m_{10}^n M_1^n}{(M_3^n)^2} \\ \frac{m_4^n}{M_3^n} - \frac{m_8^n M_1^n}{(M_3^n)^2} & \frac{m_2^n}{M_3^n} - \frac{m_{10}^n M_1^n}{(M_3^n)^2} & \frac{m_6^n}{M_3^n} - \frac{m_{10}^n M_1^n}{(M_3^n)^2} \end{bmatrix}_{2n\times 3}, \tag{10}$$

$$d\boldsymbol{X} = \begin{bmatrix} X - X_0 \\ Y - Y_0 \\ Z - Z_0 \end{bmatrix}_{3\times 1}. \tag{11}$$

The shift term of Eq. (9) is:

$$\boldsymbol{\psi} - F(\boldsymbol{X}_0) = \boldsymbol{H}d\boldsymbol{X} + \boldsymbol{\Delta}. \tag{12}$$

Let $d\boldsymbol{\psi} = \boldsymbol{\psi} - F(\boldsymbol{X}_0)$, then:

$$d\boldsymbol{X} = (\boldsymbol{H}^T \boldsymbol{H})^{-1} \boldsymbol{H}^T d\boldsymbol{\psi}. \tag{13}$$

The modified $\boldsymbol{X}_0$ is:

$$\boldsymbol{X}_0 = \boldsymbol{X}_0 + d\boldsymbol{X}. \tag{14}$$

Select the appropriate iterative threshold $\Delta_{\text{threshold}}$. When the correction value satisfies the following formula, the iterative process is exited, and the obtained $\boldsymbol{X}_0$ is the final estimation result:

$$\|d\boldsymbol{X}\| \le \Delta_{\text{threshold}}. \tag{15}$$

### *2.2 Least squares algorithm based on laser ranging*

Similarly, the least square solution $(x, y, z)$ based on ranging information can be obtained.

When locating based on distance information, the laser ranging observation can be expressed as:

$$d_i = \sqrt{(x - x_i)^2 + (y - y_i)^2 + (z - z_i)^2}, \tag{16}$$

where $(x_i, y_i, z_i)$ is the position of the flying platform at the time of the $i$-th observation, $d_i$ is the laser ranging value, and $(x, y, z)$ is the three-dimensional coordinate of the target, then the ranging error model can be expressed as:

$$\boldsymbol{\varphi} = f(\boldsymbol{x}) + \boldsymbol{\delta}, \tag{17}$$

where $\boldsymbol{\varphi}$ is the observed value, $\boldsymbol{\delta}$ is the observation error, and $f(\boldsymbol{x})$ is the nonlinear function of the independent variable $\boldsymbol{x} = [x \quad y \quad z]^T$:

$$\boldsymbol{\varphi}_{n\times 1} = \begin{bmatrix} d_1 & \cdots & d_n \end{bmatrix}^T, \tag{18}$$

$$f(\boldsymbol{x}) = \begin{bmatrix} \sqrt{(x - x_1)^2 + (y - y_1)^2 + (z - z_1)^2} \\ \vdots \\ \sqrt{(x - x_n)^2 + (y - y_n)^2 + (z - z_n)^2} \end{bmatrix}_{n\times 1}, \tag{19}$$

$$\boldsymbol{\delta}_{n\times 1} = \begin{bmatrix} \delta_{d_1} & \cdots & \delta_{d_n} \end{bmatrix}^T. \tag{20}$$

Similarly, we use the result of the forward intersection of $n$ frames of images [27] as the initial value $\boldsymbol{x}_0$, and carry out the first-order Taylor expansion of $f(\boldsymbol{x})$ at $\boldsymbol{x}_0$, then:

$$f(\boldsymbol{x}) = f(\boldsymbol{x}_0) + \boldsymbol{h} d\boldsymbol{x} + \boldsymbol{\delta}, \tag{21}$$

where $\boldsymbol{h}$ is the design matrix, $d\boldsymbol{x}$ is the correction value of the initial value $\boldsymbol{x}_0$:

$$\boldsymbol{h}_{n\times 3} = \begin{bmatrix} \frac{x_0 - x_1}{d_1} & \frac{y_0 - y_1}{d_1} & \frac{z_0 - z_1}{d_1} \\ \vdots & \vdots & \vdots \\ \frac{x_0 - x_n}{d_n} & \frac{y_0 - y_n}{d_n} & \frac{z_0 - z_n}{d_n} \end{bmatrix}, \tag{22}$$

$$d\boldsymbol{x} = \begin{bmatrix} x - x_0 \\ y - y_0 \\ z - z_0 \end{bmatrix}_{3\times 1}. \tag{23}$$

The shift term of Eq. (21) is:

$$\boldsymbol{\varphi} - f(\boldsymbol{X}_0) = \boldsymbol{h} d\boldsymbol{x} + \boldsymbol{\delta}. \tag{24}$$

Let $d\boldsymbol{\varphi} = \boldsymbol{\varphi} - f(\boldsymbol{x}_0)$, then:

$$d\boldsymbol{x} = (\boldsymbol{h}^T \boldsymbol{h})^{-1} \boldsymbol{h}^T d\boldsymbol{\varphi}. \tag{25}$$

The modified $\boldsymbol{x}_0$ is:

$$\boldsymbol{X}_0 = \boldsymbol{X}_0 + d\boldsymbol{X}. \tag{26}$$

Select the appropriate iterative threshold $\delta_{\text{threshold}}$. When the correction value satisfies the following formula, the iterative process is exited, and the obtained $x_0$ is the final estimation result:

$$\|d\boldsymbol{x}\| \leq \delta_{\text{threshold}}. \tag{27}$$

### *2.3 Least squares algorithm based on visual and laser ranging*

When the flight platform flies to the $i$-th track point, the visual observation is the pixel coordinate $(\tilde{x}_i, \tilde{y}_i)$, and the ranging observation is the laser ranging value $d_i$. We can define a new observation quantity as:

$$\boldsymbol{\Phi}_{3n\times1} = \begin{bmatrix} \boldsymbol{\psi}_{2n\times1} \\ \boldsymbol{\varphi}_{n\times1} \end{bmatrix}. \tag{28}$$

Then, the nonlinear function of vision and distance measurement regarding the independent variable $X$ can be expressed as:

$$\mathcal{L}(\boldsymbol{X})_{3n\times1} = \begin{bmatrix} F(\boldsymbol{X})_{2n\times1} \\ f(\boldsymbol{X})_{n\times1} \end{bmatrix}. \tag{29}$$

Combining Eq. (4) and Eq. (17), the new observations satisfy:

$$\boldsymbol{\Phi} = \mathcal{L}(\boldsymbol{X}) + \boldsymbol{\Lambda}, \tag{30}$$

where $\boldsymbol{\Lambda}$ is the new observation error:

$$\boldsymbol{\Lambda}_{3n\times1} = \begin{bmatrix} \boldsymbol{\Delta}_{2n\times1} \\ \boldsymbol{\delta}_{n\times1} \end{bmatrix}. \tag{31}$$

Combining Eq. (9) and Eq. (21), the Taylor first-order expansion of $\mathcal{L}(\boldsymbol{X})$ at $\boldsymbol{X}_0 = [X_0 \quad Y_0 \quad Z_0]^T$ is carried out, and the remainder is discarded:

$$\mathcal{L}(\boldsymbol{X}) = \mathcal{L}(\boldsymbol{X}_0) + \boldsymbol{\Gamma} d\boldsymbol{X} + \boldsymbol{\Lambda}, \tag{32}$$

where $\boldsymbol{\Gamma}$ is the design matrix, and $d\boldsymbol{X}$ is the correction value of the target initial value $\boldsymbol{X}_0$:

$$\boldsymbol{\Gamma}_{3n\times3} = \begin{bmatrix} \boldsymbol{H}_{2n\times3} \\ \boldsymbol{h}_{n\times3} \end{bmatrix}, \tag{33}$$

$$d\boldsymbol{X}_{3\times1} = \begin{bmatrix} X - X_0 \\ Y - Y_0 \\ Z - Z_0 \end{bmatrix}. \tag{34}$$

The shift term of Eq. (32) is:

$$\boldsymbol{\Phi} - \mathcal{L}(\boldsymbol{X}_0) = \boldsymbol{\Gamma} d\boldsymbol{X} + \boldsymbol{\Lambda}. \tag{35}$$

Considering that the physical units of visual and ranging information are different, it is necessary to normalize the data. We regard $d\boldsymbol{\psi} = \boldsymbol{\psi} - F(\boldsymbol{X}_0)$ and $d\boldsymbol{\varphi} = \boldsymbol{\varphi} - f(\boldsymbol{X}_0)$ as a whole for normalization, respectively. $Max$ and $Min$ are the maximum and minimum values of $d\psi$ , and $max$ and $min$ are the maximum and minimum values of $d\varphi$ . According to Eq. (35), we can get:

$$\begin{bmatrix} \dfrac{d\psi - Min}{Max - Min} \\ \dfrac{d\varphi - min}{max - min} \end{bmatrix} = \begin{bmatrix} \dfrac{1}{Max - Min}\boldsymbol{H} & \dfrac{-Min}{Max - Min} \\ \dfrac{1}{max - min}\boldsymbol{h} & \dfrac{-min}{max - min} \end{bmatrix} \begin{bmatrix} dX \\ 1 \end{bmatrix} + \begin{bmatrix} \dfrac{\boldsymbol{\Delta} - Min}{Max - Min} \\ \dfrac{\boldsymbol{\delta} - min}{max - min} \end{bmatrix}. \tag{36}$$

Let:

$$d\boldsymbol{\Phi}_{3n\times 1} = \begin{bmatrix} \dfrac{d\psi - Min}{Max - Min} \\ \dfrac{d\varphi - min}{max - min} \end{bmatrix}, \tag{37}$$

$$\boldsymbol{T}_{3\times 4} = \begin{bmatrix} \dfrac{1}{Max - Min}\boldsymbol{H} & \dfrac{-Min}{Max - Min} \\ \dfrac{1}{max - min}\boldsymbol{h} & \dfrac{-min}{max - min} \end{bmatrix}. \tag{38}$$

We have:

$$\begin{bmatrix} d\boldsymbol{X} \\ 1 \end{bmatrix} = (\boldsymbol{T}^T\boldsymbol{T})^{-1}\boldsymbol{T}^T d\boldsymbol{\Phi}. \tag{39}$$

Thus, the modified $\boldsymbol{X}_0$ is:

$$\boldsymbol{X}_0 = \boldsymbol{X}_0 + d\boldsymbol{X}. \tag{40}$$

Select the appropriate iterative threshold $\Lambda_{\text{threshold}}$ , when the correction value satisfies the following formula, the iterative process is exited, and the obtained $\boldsymbol{X}_0$ is the final estimation result:

$$\|d\boldsymbol{X}\| \le \Lambda_{\text{threshold}}. \tag{41}$$

### *2.4 Fusion algorithm based on ridge estimation*

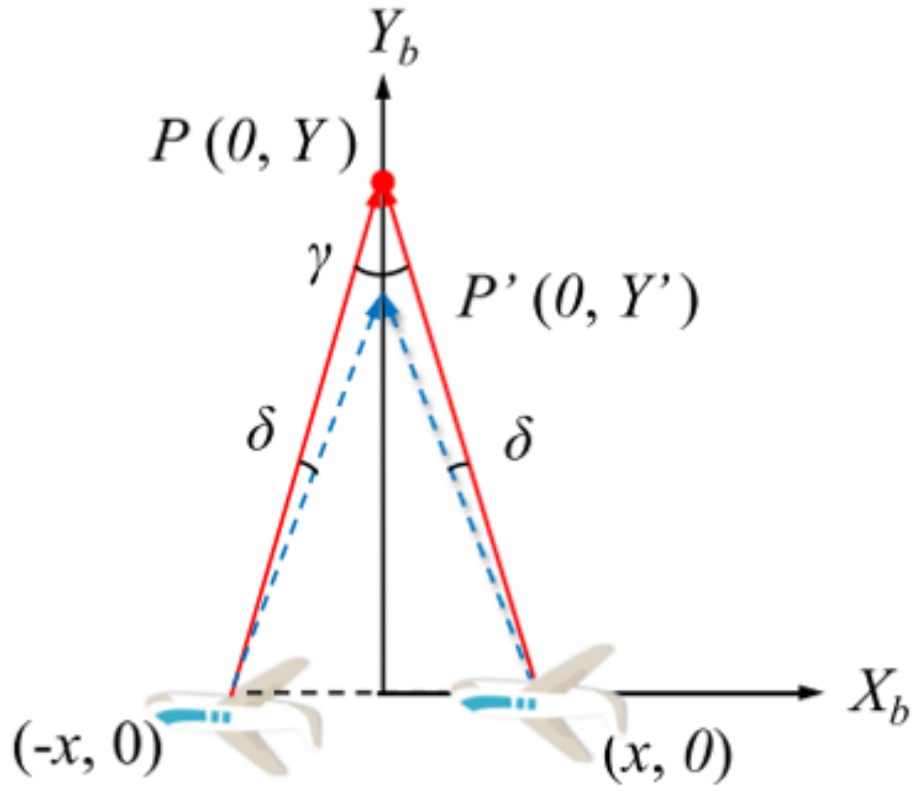

Fig. 3. The localization error caused by small intersection angles.

In actual measurements, there are limited observation conditions such as long distance, small intersection angles, and large inclination angles. As shown in Fig.3, let $P(\mathbf{0}, Y)$ be the ground truth of the target and $P'(\mathbf{0}, Y')$ be the intersection results. $\delta$ is the line-of-sight angular error. It can be derived from geometric relationships:

$$\begin{cases} \tan(\frac{\gamma}{\mathbf{2}}) = \frac{x}{Y} \\ \tan(\frac{\gamma}{\mathbf{2}} - \delta) = \frac{x}{Y'} \end{cases} \tag{42}$$

The localization error caused by small intersection angles is $(Y - Y')$. From Eq. (42), it can be seen that for the same line-of-sight angular error, the localization error increases sharply as the intersection angle decreases. Moreover, as Fig.2, a large inclination angle will result in a greater line-of-sight angular error. Under the combined influence of limited observation conditions, the design matrix of the least squares algorithm exhibits severe ill-conditioning [26]. That is, there is serious multicollinearity between column vectors, i.e. there is an approximate linear correlation between the variables. As a result, a small perturbation in the column vector will cause a significant error in the solution vector, which will lead to low accuracy and unstable results. In this case, the least squares estimation is no longer a good estimate.

Ridge estimation is an improved least squares estimation method. It improves the stability of the estimates by sacrificing some precision to reduce the mean squared error, while giving up the unbiasedness of least squares estimation. Ridge estimation is widely applied in data analysis across fields such as economics, engineering, and biomedicine. It performs very well in handling data with multicollinearity. Given the above problems, this paper introduces ridge estimation [31] and adds a penalty function to the least squares estimation objective function. By appropriately reducing the accuracy, giving up some information, and sacrificing the cost of least squares unbiasedness, more robust and more realistic regression coefficients are obtained to solve the problem of serious collinearity of design matrix column vectors.

After adding perturbation to the diagonal of the normal matrix of Eq. (39), the ridge estimation can be expressed as:

$$\begin{bmatrix} d\boldsymbol{X} \\ 1 \end{bmatrix} = (\boldsymbol{T}^T\boldsymbol{T} + k\boldsymbol{I})^{-1}\boldsymbol{T}^T d\boldsymbol{\Phi}, \tag{43}$$

where $k$ is the ridge parameter, and $\boldsymbol{I}$ is the unit matrix of size $4 \times 4$. It can be seen that the ridge estimation is a linear transformation of the least squares estimation. Under the mean square error criterion, there is $k > 0$ such that the ridge estimation is better than the least squares estimation.

Selecting an appropriate ridge parameter is crucial for ridge regression. There are various methods [32–34] to select the ridge parameters. This paper adopts the Hoerl-Kennard-Baldwin ridge parameter selection method [35]. It is an efficient approach that does not require fitting or iteration. In practice, this method demonstrates good performance. We choose this method after considering both computational efficiency and accuracy. The calculation of the ridge parameter is as follows:

$$k = \frac{t\delta_0^2}{\hat{\boldsymbol{X}}^T\boldsymbol{T}^T\boldsymbol{T}\hat{\boldsymbol{X}}}, \tag{44}$$

where $t$ is the rank of matrix $\boldsymbol{T}$, $\hat{X}$ is the value of the least squares estimate, and $\delta_0^2$ is :

$$\delta_0^2 = \frac{\boldsymbol{\Phi}^T\left[\boldsymbol{I}-\boldsymbol{T}(\boldsymbol{T}^T\boldsymbol{T})^{-1}\boldsymbol{T}^T\right]\boldsymbol{\Phi}}{n-t}. \tag{45}$$

Thus, the modified $\boldsymbol{X}_0$ is:

$$\boldsymbol{X}_0 = \boldsymbol{X}_0 + d\boldsymbol{X}. \tag{46}$$

Select the appropriate iterative threshold $\Lambda_{\text{threshold}}$, when the correction value satisfies the following formula, the iterative process is exited, and the obtained $\boldsymbol{X}_0$ is the final estimation result:

$$\|d\boldsymbol{X}\| \le \Lambda_{\text{threshold}}. \tag{47}$$

The specific process of the localization algorithm proposed in this paper is shown in Table 1.

**Table 1 Algorithm flow.**

| Algorithm | Vision and laser fusion localization algorithm based on ridge estimation |
|---|---|
| Application： | High altitude, long distance, large inclination, small intersection angle, and other restricted observation conditions |
| Input： | $\boldsymbol{K}$ - internal parameter matrix,<br>**M** - projection matrix,<br>$\boldsymbol{d}$ - laser ranging,<br>Locations of the flight platform,<br>n images. |
| Output: | $[X \quad Y \quad Z]^T$ - target location. |
| 1: | The visual localization or laser ranging localization result is used as the initial value $X_0$ of the least squares algorithm iteration. |
| 2: | Establish the observation equation based on visual and laser fusion as shown in Eq. (30). |
| 3: | Carry out Taylor expansion of $\mathcal{L}(\boldsymbol{X})$ at $X_0$ to obtain the design matrix $\boldsymbol{\Gamma}$. |
| 4: | Normalize the vision and ranging data. |
| 5: | Calculate the ridge parameter $k$. |
| 6: | Select an appropriate iteration threshold $\Lambda_{\text{threshold}}$ and iteratively solve the target's location. |

## 3. Experiments

### 3.1 Simulation experiments

The flight platform is equipped with a combined positioning and navigation system, a monocular camera, a laser rangefinder, etc., and 10 continuous observations are performed on the ground fixed target at a height of 2 km and a distance of 5 km. The observation angle $\theta$ is 66.42°. In this paper, a variety of noises satisfying the normal distribution with a mean of zero are introduced, including the position noise of the flight platform with a standard deviation of 5m (random noise is added in the three-axis direction), the laser ranging noise with a standard deviation of 2.5m, the relative rotation angle noise with a standard deviation of 0.2 °, and the pixel extraction noise with a standard deviation of 0.1 pixel. A thousand independent simulation experiments were carried out when the intersection angle $\gamma$ of the head and tail track points were 10 °, 20 °,30 °, 40 °, 50 °, 60 °, 70 °, and 80 ° respectively, and the median error of multiple experiments were taken.

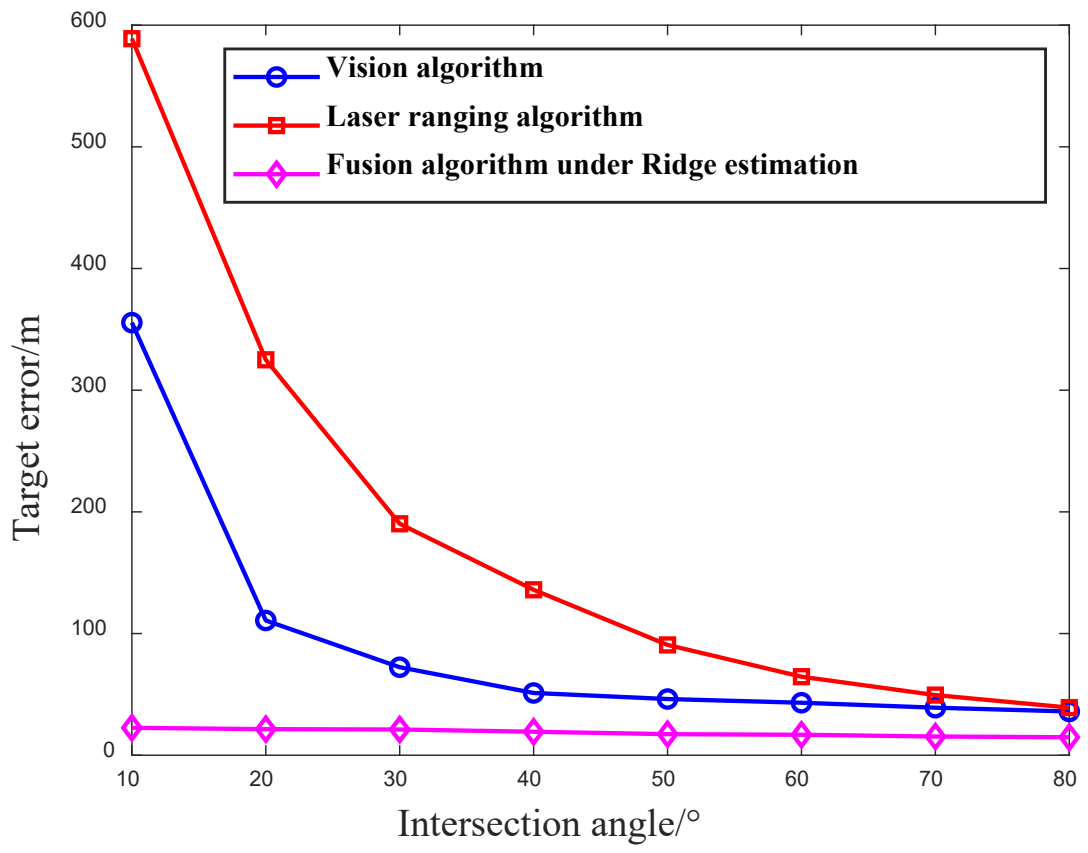

Fig. 4. Ground target localization error of three different algorithms.

In this paper, the iterative threshold $\Lambda_{threshold}$ is 0.0001, and the result of the visual localization algorithm [27] is used as the initial value of the least squares iteration of the fusion algorithm. The localization error of the four different algorithms is shown in Table 2. The localization error of the vision-based algorithm, the laser ranging-based algorithm, and the fusion algorithm under the ridge estimation proposed in this paper are drawn in Fig.4.

**Table 2 Localization error of four different algorithms.**

| Intersection angle /° | 10 | 20 | 30 | 40 | 50 | 60 | 70 | 80 |
|---|---|---|---|---|---|---|---|---|
| Vision algorithm [27]/m | 355.52 | 110.74 | 72.33 | 51.17 | 46.19 | 43.14 | 39.06 | 35.96 |
| Laser ranging algorithm [36]/m | 588.89 | 324.99 | 190.20 | 135.97 | 90.66 | 64.56 | 49.44 | 39.14 |
| Fusion algorithm/m | 23.39 | 22.44 | 21.89 | 20.00 | 18.15 | 17.18 | 15.85 | 15.10 |
| Fusion algorithm under Ridge estimation/m | 22.52 | 21.43 | 21.14 | 19.30 | 17.32 | 16.81 | 15.39 | 14.79 |

It can be seen from Table 2 that as the intersection angle increases, the observation conditions become better, and the localization accuracy of the visual algorithm and the laser ranging algorithm gradually increases. When the intersection angle is less than 30 degrees, as the intersection angle decreases, the error increases sharply. This may be because when the intersection angle is less than 30 degrees, there is serious multicollinearity between column vectors. As a result, the algorithm using a single sensor is significantly affected by observation noises, leading to low localization accuracy. In the case of fewer track points, the visual information is more abundant, and the vision-based localization algorithm has higher localization accuracy than the laser-ranging-based algorithm. Moreover, the localization accuracy of the fusion visual and laser ranging algorithm is higher than that of the algorithm using certain information alone. Especially when the intersection angle is relatively small, the improvement in the accuracy of the fusion algorithm is quite significant.

After introducing ridge estimation, the observation conditions are improved, and the localization accuracy of the algorithm is further improved. However, there is not much difference in positioning error between the fusion algorithm and the fusion algorithm under ridge estimation. This may be because the data for the fusion algorithm come from two different sensors. Therefore, even if there is a strong linear correlation within each kind of data, there is no linear correlation between the two kinds of data. This fusion can mitigate the multicollinearity of the design matrix. To validate the stability and robustness of the ridge estimation method, we employed root mean square (RMS) error as an indicator.

When the design matrix exhibits ill-conditioning under limited observation conditions, the traditional least squares estimation often results in a significant RMS error, leading to low estimation accuracy and instability. As a consequence, least squares estimation ceases to be a desirable choice in such scenarios. Therefore, to mitigate the ill-conditioning of the design matrix and reduce the measurement RMS error, this paper introduced ridge estimation.

Under the simulation conditions and noise conditions of the aforementioned experiments, consecutive observations were made 10 times each for intersection angles of 10 °, 20 °,30 °, 40 °, 50 °, 60 °, 70 °, and 80 ° formed by the head and tail track points. After a thousand independent simulation experiments were carried out, we calculated the RMS error values of the fusion algorithm and the fusion algorithm under ridge estimation. The results are presented in Fig.5.

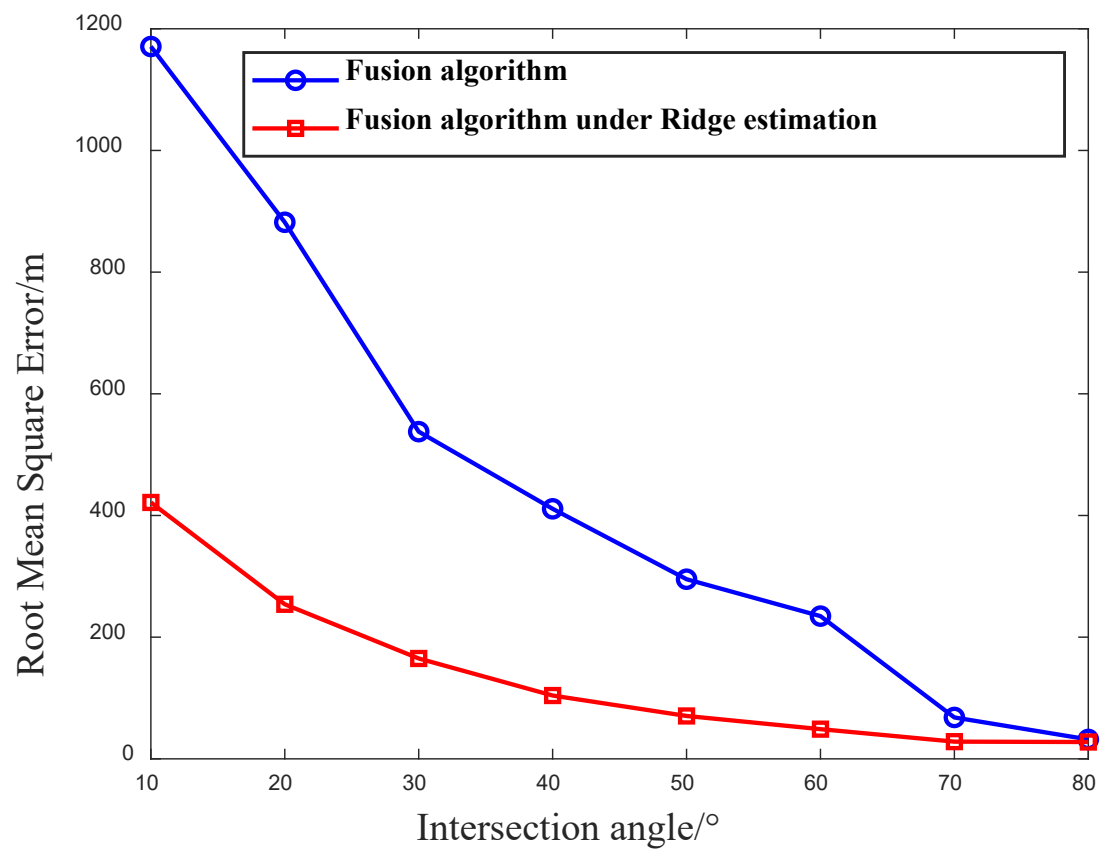

Fig. 5. RMS Error of the fusion algorithm and the fusion algorithm under ridge estimation.

It can be seen from Table 2 and Fig.5 that compared to the fusion algorithm, although the introduction of ridge estimation only slightly improves the measurement accuracy, it significantly reduces the RMS error of the measurements, particularly under limited observational conditions. Therefore, the fusion algorithm under ridge estimation proposed in this paper exhibits strong stability and high robustness under limited observation conditions.

Ridge estimation is an improved least squares estimation method. It is essentially a linear transformation of the least squares method, which solves linear equations. Therefore, the fusion algorithm under ridge estimation proposed in this paper has extremely high computational efficiency. Simulation experiments were conducted on a 12th Gen Intel(R) Core(TM) i9-12900H CPU with an intersection angle of 30° and 10 observations, resulting in a runtime of around 3.528 s. Most of the time is spent on the vision algorithm, as it requires feature matching and computation of camera pose and its results are required to calculate the ridge parameters. The time required for the fusion algorithm under ridge estimation is only around 0.001 s. The results demonstrate that the proposed algorithm exhibits high computational efficiency, almost achieving simultaneous output with the vision algorithm.

### *3.2 Flight experiments*

In this section, we use six sets of flight experimental data for localization analysis. The flight platform is equipped with a monocular camera and a laser range finder to localization the target, and the internal and external parameters of the observation equipment are precisely known. The flight platform tracks and locks the square target with a cross shape in the center of the field of view, performing visual shooting and laser ranging simultaneously, as shown in Fig.6. The monocular camera captures images at a spatial resolution of 600 × 400 pixels and a temporal resolution of 25 Hz. The laser ranging error is around 5 m. The position error of the flight platform is around 2 m. The attitude error of the flight platform is around 0.2°. The observation of the flight platform is about 5 km, and the observation angle is about 66.42°.

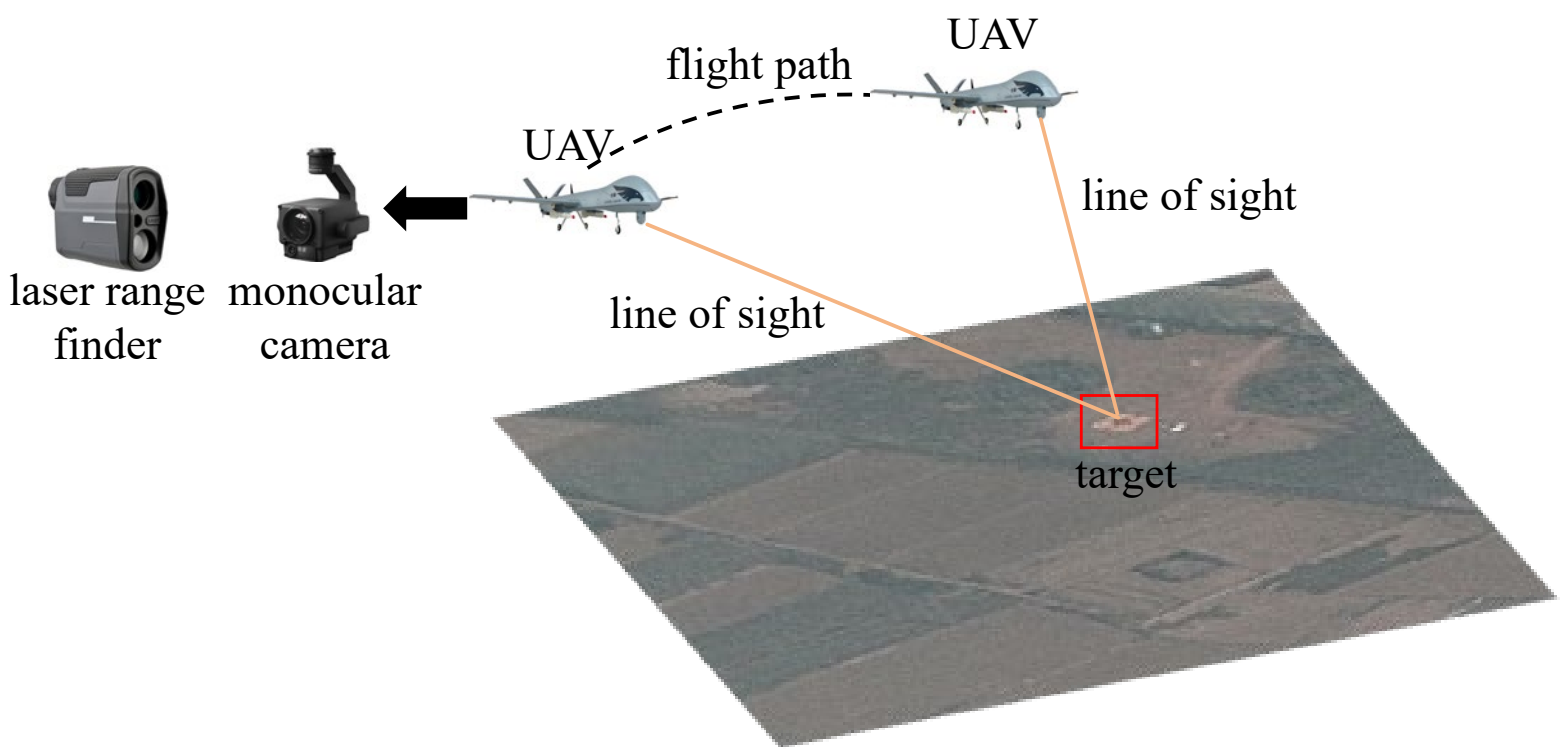

Fig. 6. Equipped with a monocular camera and a laser range finder, UAV tracks and localizes the target.

**Table 3 Flight parameters.**

| Group | 1 | 2 | 3 | 4 | 5 | 6 |
|---|---|---|---|---|---|---|
| Intersection angle /° | 53.97 | 24.03 | 24.36 | 31.42 | 46.46 | 36.41 |
| Observation times/time | 16 | 17 | 34 | 11 | 18 | 11 |

The illustrations of the flight path, the observation line of sight (LOS), the ground truth of the targets and the localization results calculated by three different algorithms are shown in Fig.7. The intersection angles and observation times of the six sets of data are shown in Table 3. And the localization errors of the six sets of data that are calculated by three different algorithms are shown in Table 4.

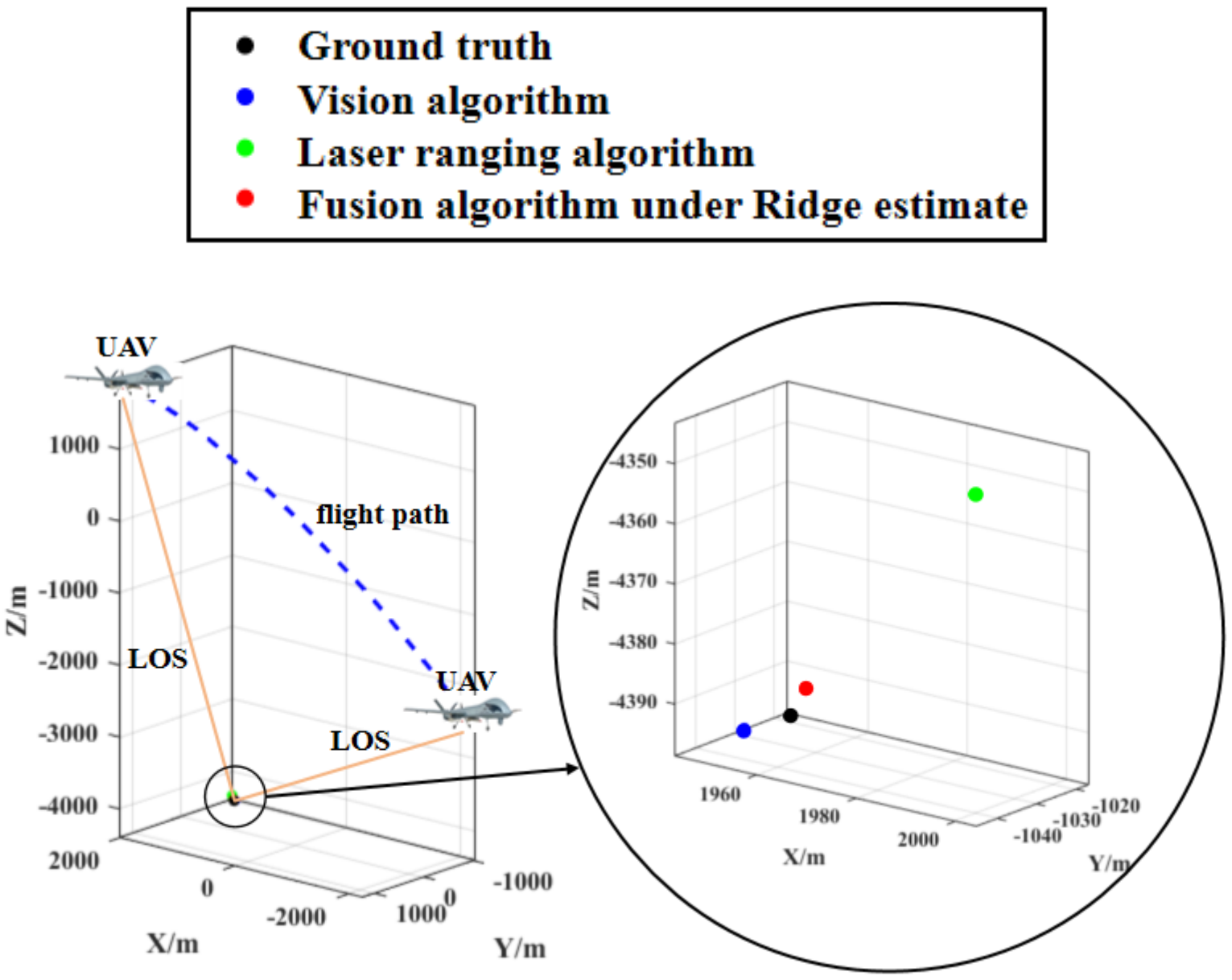

(a) Illustrations of group 1.

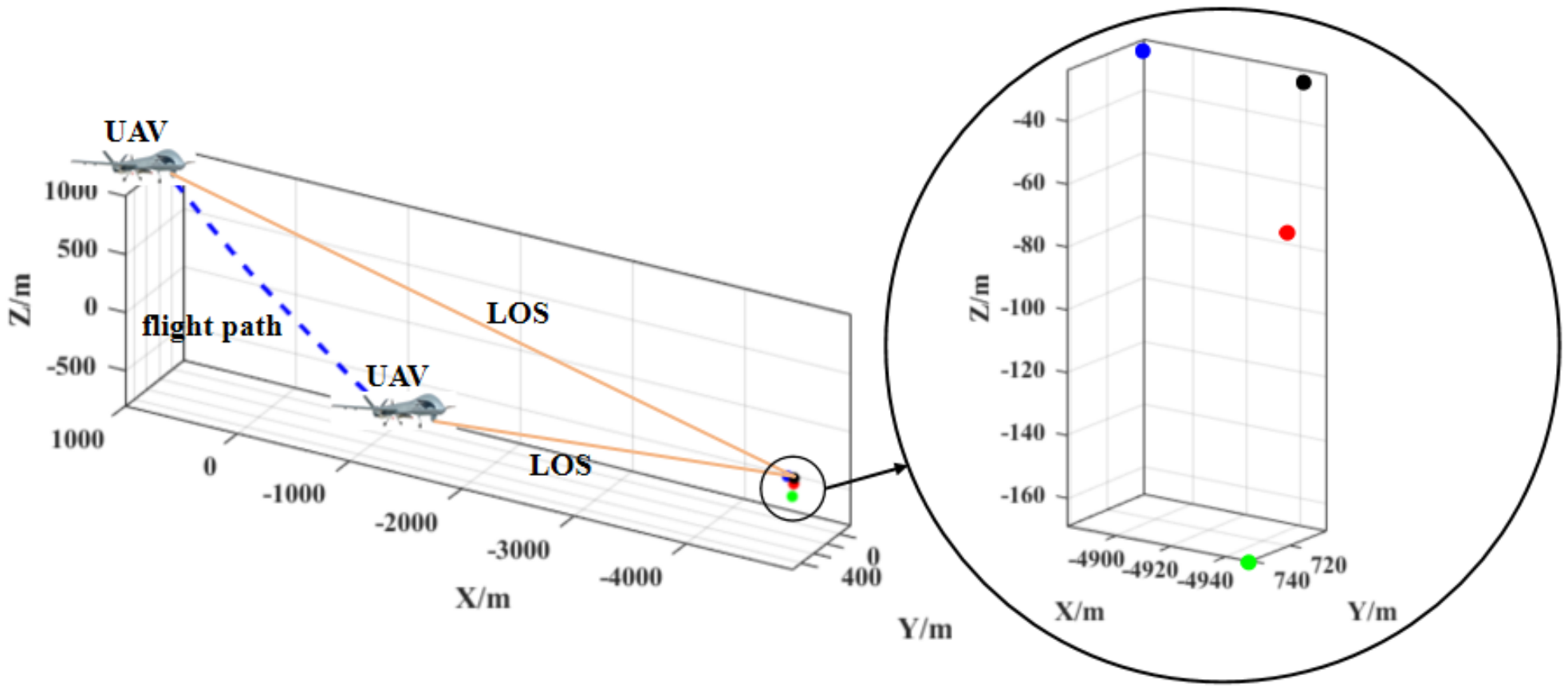

(b) Illustrations of group 2.

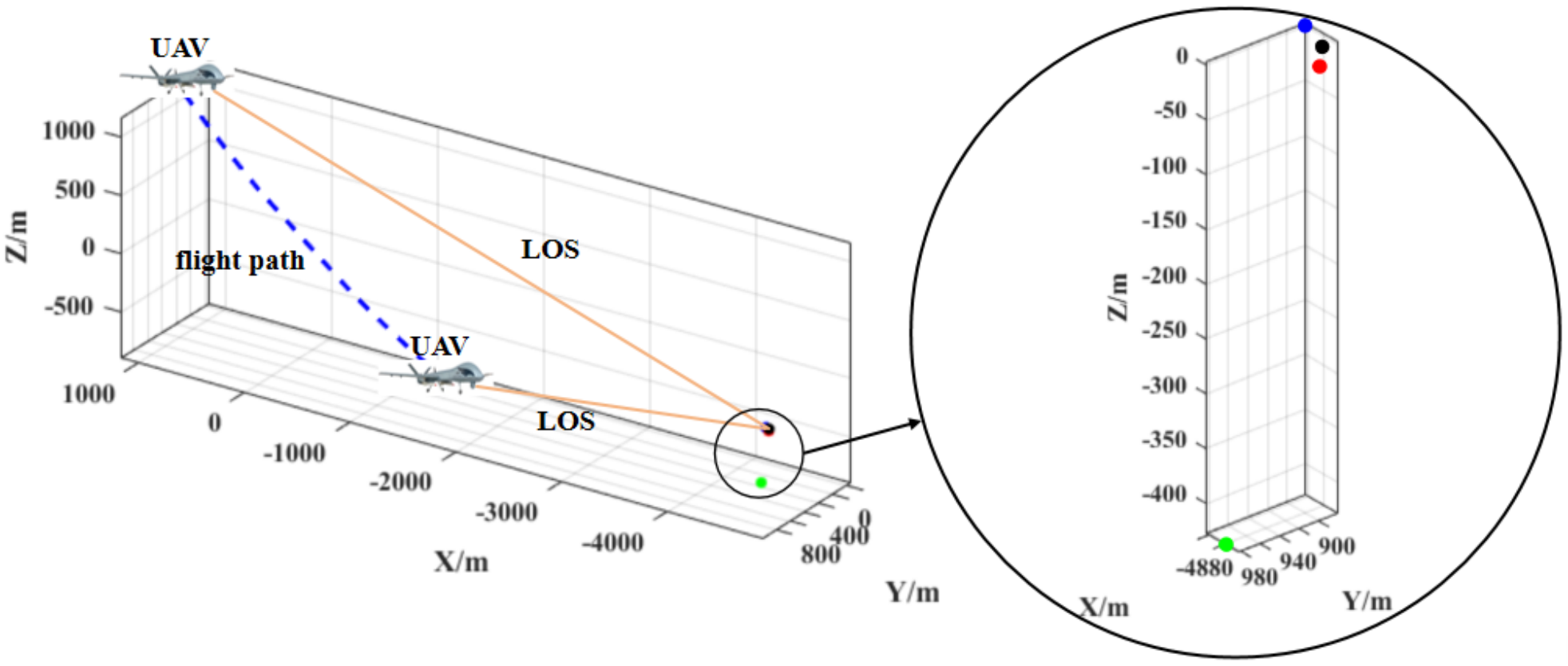

(c) Illustrations of group 3.

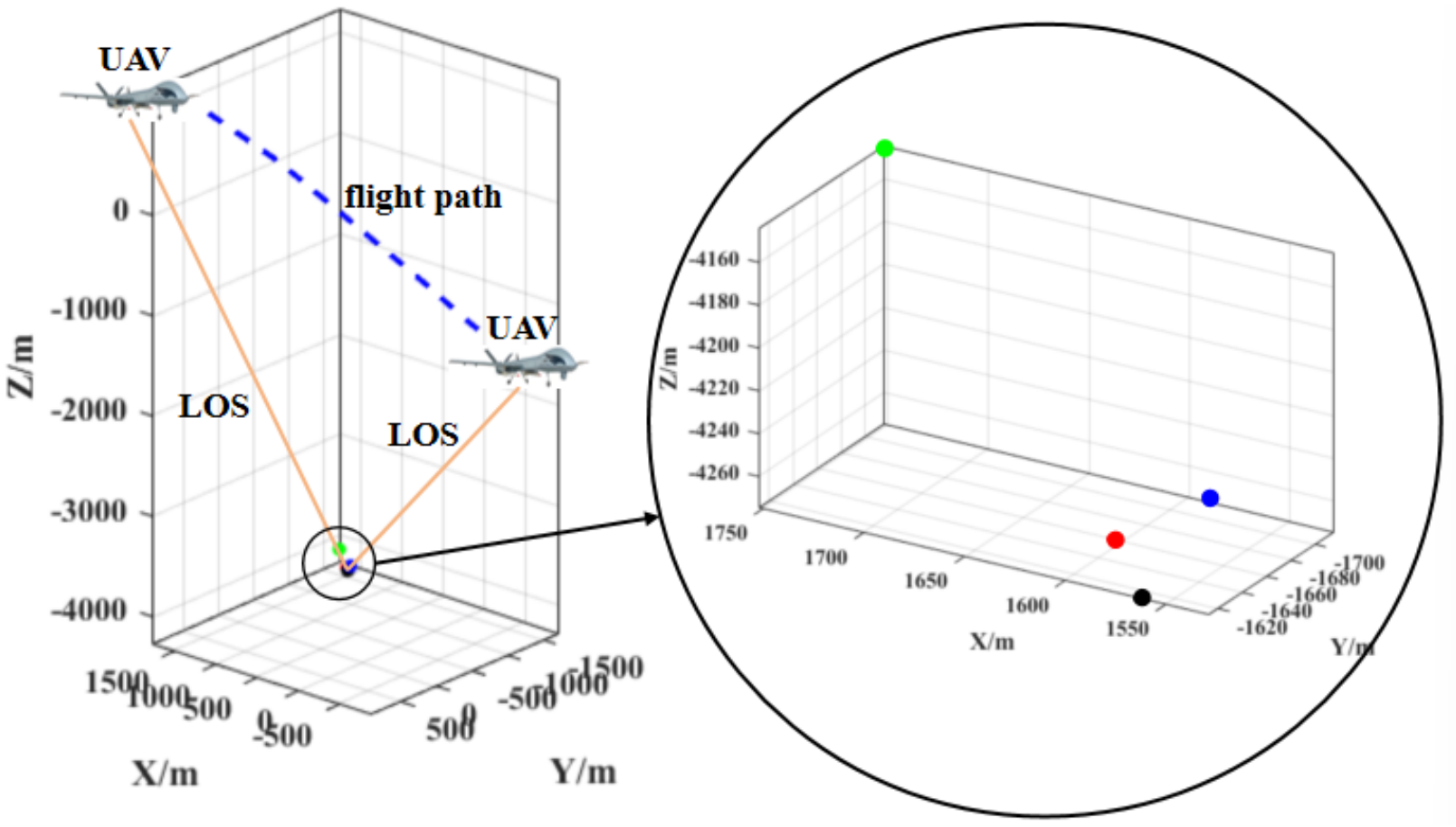

(d) Illustrations of group 4.

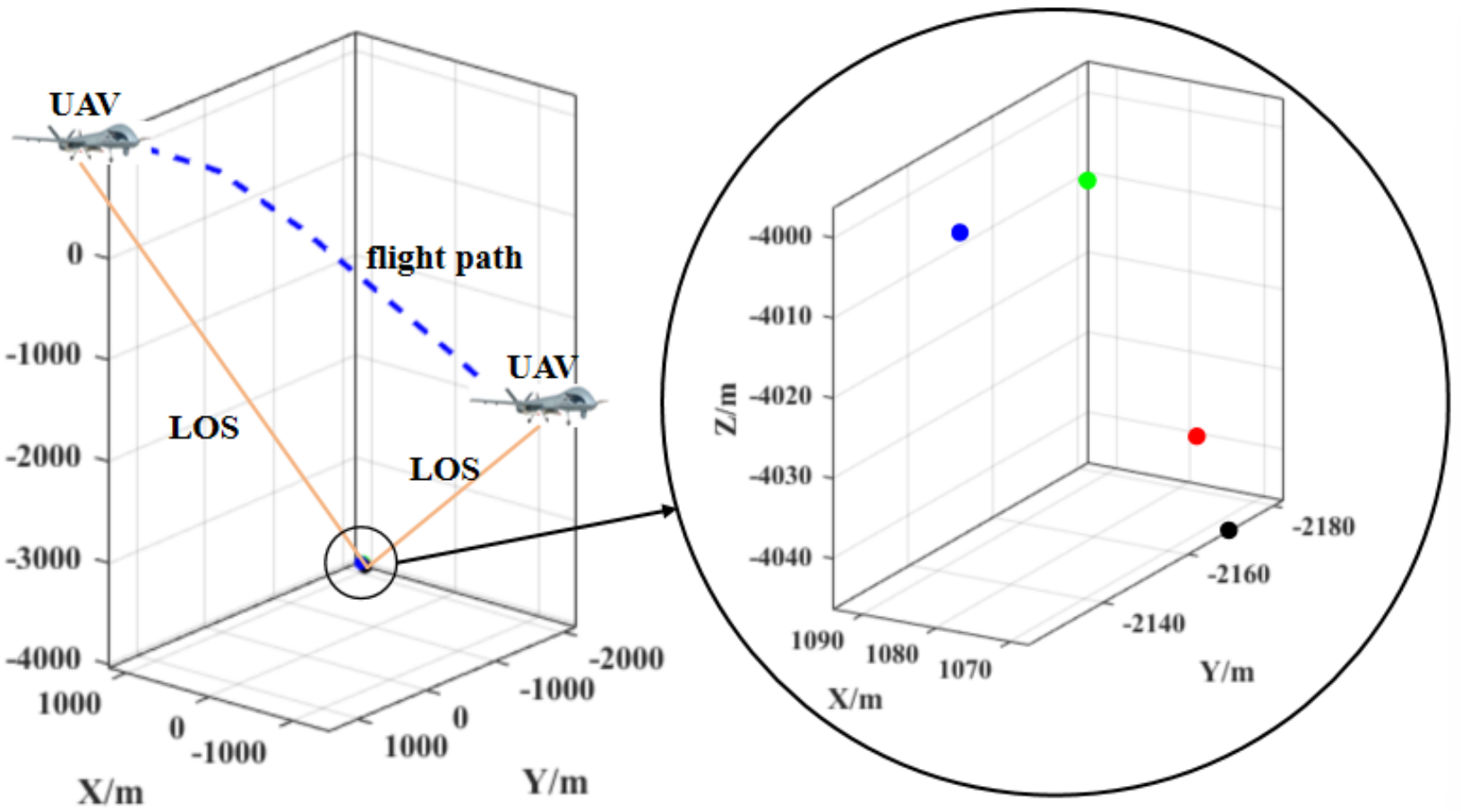

(e) Illustrations of group 5.

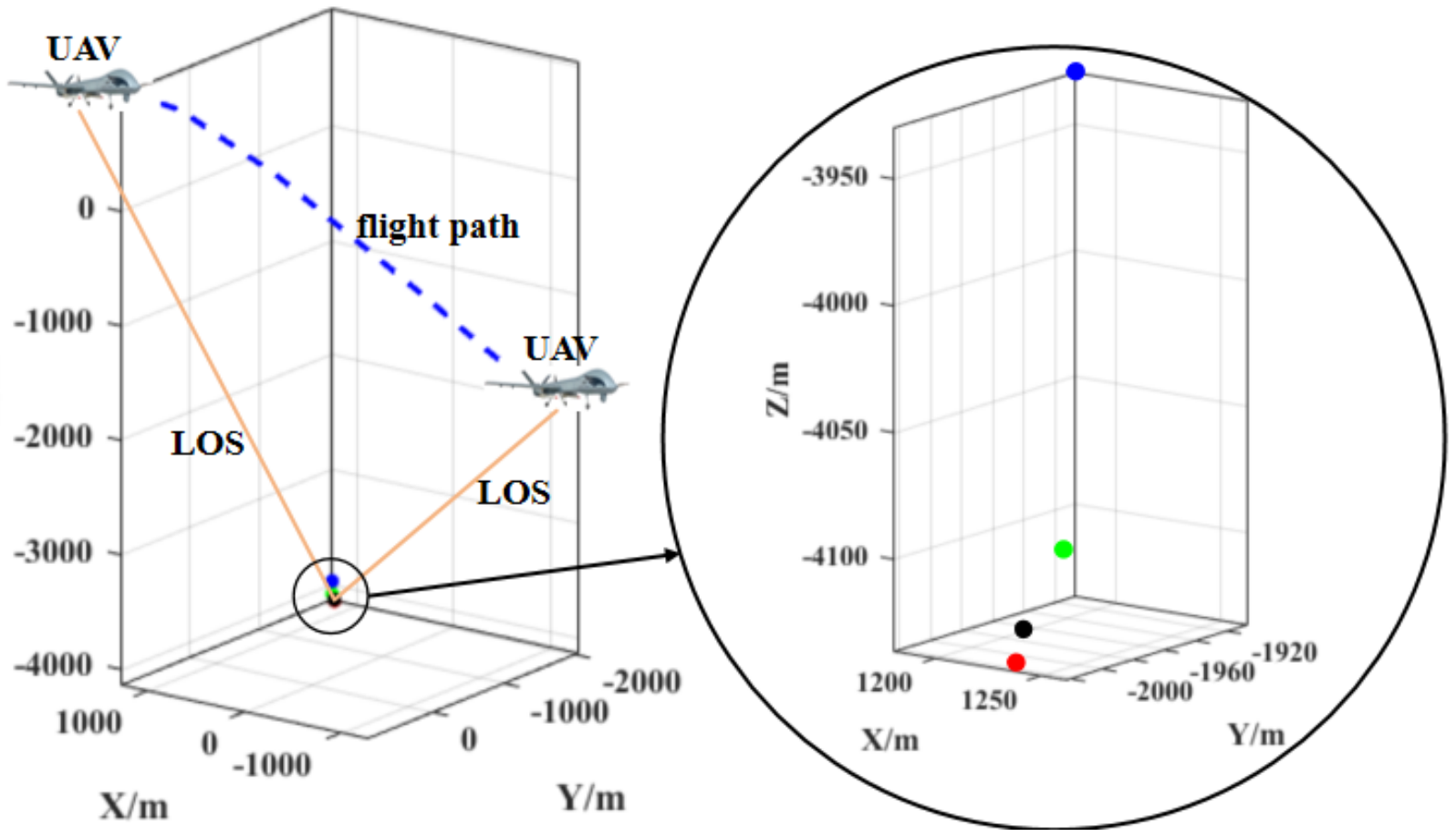

(f) Illustrations of group 6.

Fig. 7. The illustrations of six sets of flight experiment and localization results of three different algorithms.

As shown in Fig.6 and Fig.7, the flight platform is equipped with a monocular camera and a laser range finder, performing visual shooting and laser ranging simultaneously. We calculate the target's position using three different algorithms, which are the vision algorithm, the laser ranging algorithm, the fusion algorithm without normalization, the fusion algorithm, and the fusion algorithm under ridge estimation, respectively. The distance information obtained from the laser range finder can be used directly in the laser-ranging-based algorithm and the fusion algorithm under ridge estimation. However, the images captured by the monocular camera cannot directly be used. Instead, it requires extracting feature points from the images and performing feature point matching. The fusion algorithms and the vision-based algorithm employ the same feature point extraction and matching algorithm, the SIFT algorithm.

The flight experiment results in Fig.7 and Table 4 are consistent with the simulation results. It can be seen easily from Table 4 that when processing the first, second, third, and fourth data groups, the vision-based algorithm has higher localization accuracy than the laser-ranging-based algorithm. And when dealing with the fifth and sixth data groups, the laser-ranging-based algorithm has higher localization accuracy than the vision-based algorithm. This is because the localization accuracy of vision-based algorithms is not only influenced by factors such as observation conditions, number of observations, and intersection angles, but also closely related to the precision of feature point extraction and matching. When processing the second, third, and fourth data groups, due to the small intersection angle, the laser-ranging-based algorithm exhibits significant positioning errors. This result is consistent with the simulation results under limited observation conditions.

**Table 4 Localization error of six groups of data.**

| Group | 1 | 2 | 3 | 4 | 5 | 6 |
|---|---|---|---|---|---|---|
| Vision algorithm [27]/m | 10.75 | 68.12 | 42.87 | 65.27 | 69.34 | 223.44 |
| Laser ranging algorithm [36]/m | 86.42 | 148.09 | 439.99 | 250.92 | 45.41 | 56.04 |
| Line of sight algorithm [37]/m | 175.17 | 24252.97 | 24283.62 | 201.30 | 198.18 | 87.53 |
| Fusion algorithm without normalization/m | 10.16 | 62.53 | 42.68 | 49.48 | 12.17 | 26.02 |
| Fusion algorithm/m | 9.03 | 58.86 | 29.47 | 44.51 | 12.10 | 14.56 |
| Fusion algorithm under Ridge estimation/m | 8.68 | 46.33 | 17.32 | 28.43 | 11.48 | 13.73 |

However, it can be seen from Fig.7 and Table 4 that no matter what kind of information the algorithm is based on has higher localization accuracy, the fusion localization algorithm under ridge estimation has the lowest localization error. It is noteworthy that, for accurate comparison of the accuracy between the vision-based algorithm and fusion algorithm under ridge estimation, we employ identical feature point extraction and matching results in our computations. Therefore, we can conclude that the fusion algorithm under ridge estimation can achieve the best localization results under any observation conditions.

In the six sets of experiments, the intersection angles in group 2 and group 3 are less than 30 degrees, satisfying the condition of a small intersection angle. And the intersection angles in these two sets of experiments are almost the same. For both the vision algorithm and the fusion algorithm, the localization errors in group 3 are smaller than those in group 2. This is because the data of group 3 has more observation times. Instead, for the laser ranging algorithm, the localization error in group 3 is much larger. This may be because the laser ranging algorithm is more susceptible to the influence of lighting conditions. However, this do not affect the accuracy of the fusion algorithm. This result demonstrates that the fusion algorithm combines the advantages of both sensors and exhibits strong stability under limited observation conditions.

To validate the performance of the proposed fusion algorithm, we compare the proposed algorithm with the fusion algorithm employed by Ge et al [37]. Their method calculates the direction of the target through a single image frame. Then, the target's position is obtained by combining the target's direction, the platform's position, and the distance information from the laser rangefinder. This method requires only a single observation from the monocular camera and the laser rangefinder to localize the target. Since the real data includes multiple observations, we select the positioning result with the smallest localization error for comparison. This method is referred to as line of sight algorithm, as shown in Table 4. This algorithm does not fully utilize the observational data. The localization accuracy is even lower than that of the least squares method using a single sensor. Moreover, this method is highly unstable. For instance, in group 2 and group 3, it obtains completely incorrect results.

We also utilize the fusion algorithm and the fusion algorithm without normalization to locate the target, as shown in Table 4. The proposed fusion algorithm under ridge estimation still has the highest localization accuracy. Under limited observation conditions, such as group 2, group 3, and group 4, the accuracy of the fusion algorithm under ridge estimation is significantly higher than that of the fusion algorithm. This result validates that ridge estimation can mitigate the ill-conditioned problems. Moreover, the accuracy of the fusion algorithm without normalization is lower than that of the fusion algorithm. When there is a significant difference in accuracy between the vision algorithm and the laser ranging algorithm, the fusion algorithm without normalization is more greatly affected, as group 1, group 3, and group 6. Normalization is equivalent to weighting the data of two sensors appropriately. It has good performance especially when there is a significant difference in accuracy between the vision and laser ranging algorithms. This result validates the effectiveness of the fusion strategy and normalization method proposed in this paper.

The results of the above six sets of flight experiments sufficiently demonstrate that the fusion localization algorithm under ridge estimation proposed in this paper has high localization accuracy and stability.

## 4. Conclusions

Aiming at the issue of low measurement accuracy in the visual axis direction of monocular cameras and the reliance of laser-based ranging positioning methods on multiple consecutive observations, we utilize the airborne photoelectric platform to continuously observe the target, and the ground target is located by combining the monocular camera and the laser range finder. By normalizing the visual and distance data, the objective function satisfies the visual and distance constraints at the same time. It combines the rich scene information of the sequence image and the high-precision advantages of laser ranging. Ridge estimation is introduced because the design matrix of the least squares algorithm is ill-posed under limited observation conditions. A fusion localization method based on ridge estimation is proposed. To validate the superiority of the proposed algorithm in this paper, we conduct multiple simulation experiments at various intersection angles and six sets of flight experiments. Simulation and flight experiments show that the fusion method based on ridge estimation has higher localization accuracy and better robustness than the vision-based and range-based localization algorithms. Especially under the limited observation conditions, the localization accuracy of the proposed method is more than ten times higher than that of using a single sensor. In the flight experiment, with an observation distance of 6 km, the localization errors are all less than 50 m. The proposed method is applicable to various tasks of unmanned aerial vehicles. It also holds the prospects to be extended to other platforms, such as autonomous vehicles and intelligent robots.

**Funding.** This research has been supported by the Hunan Provincial Natural Science Foundation for Excellent Young Scholars under Grant 2023JJ20045, and the National Natural Science Foundation of China under Grant 12372189.

**Disclosures.** The authors declare no potential conflicts of interest.